\documentclass[11pt]{article}
\usepackage{colacl06}
\usepackage{times}
\usepackage{psfig}
\usepackage{haldefs}
\usepackage{url}
\newcommand{\pinch}{\vspace{-0.8mm}}
\newcommand{\pitem}{\vspace{-0.8mm}\item}

\title{Bayesian Query-Focused Summarization}

\newcommand{\bayesum}{\textsc{BayeSum}}

\author{Hal Daum\'e III \textnormal{and} Daniel Marcu\\
Information Sciences Institute\\
4676 Admiralty Way, Suite 1001\\
Marina del Rey, CA 90292\\
\texttt{me@hal3.name,marcu@isi.edu}\\
}

\date{}

\begin{document}

\maketitle

\begin{abstract}
We present \bayesum\ (for ``Bayesian summarization''), a model for
sentence extraction in query-focused summarization.  \bayesum\
leverages the common case in which multiple documents are relevant to
a single query.  Using these documents as reinforcement for query
terms, \bayesum\ is not afflicted by the paucity of information in
short queries.  We show that approximate inference in \bayesum\ is
possible on large data sets and results in a state-of-the-art
summarization system.  Furthermore, we show how \bayesum\ can be
understood as a justified query expansion technique in the language
modeling for IR framework.
\end{abstract}

\section{Introduction}

We describe \bayesum, an algorithm for performing query-focused
summarization in the common case that there are many relevant
documents for a given query.  Given a query and a collection of
relevant documents, our algorithm functions by asking itself the
following question: what is it about these relevant documents that
differentiates them from the \emph{non-}relevant documents?  \bayesum\
can be seen as providing a statistical formulation of this exact
question.

The key requirement of \bayesum\ is that multiple relevant documents
are known for the query in question.  This is not a severe limitation.
In two well-studied problems, it is the de-facto standard.  In
standard multidocument summarization (with or without a query), we
have access to known relevant documents for some user need.
Similarly, in the case of a web-search application, an underlying IR
engine will retrieve multiple (presumably) relevant documents for a
given query.  For both of these tasks, \bayesum\ performs well, even
when the underlying retrieval model is noisy.

The idea of leveraging known relevant documents is known as query
expansion in the information retrieval community, where it has been
shown to be successful in ad hoc retrieval tasks.  Viewed from the
perspective of IR, our work can be interpreted in two ways.  First, it
can be seen as an \emph{application} of query expansion to the
summarization task (or, in IR terminology, passage retrieval); see
\cite{liu02passage,murdock05sentence}.  Second, and more importantly,
it can be seen as a method for query expansion in a non-ad-hoc manner.
That is, \bayesum\ is a statistically justified query expansion method
in the language modeling for IR framework \cite{ponte98lm}.

\section{Bayesian Query-Focused Summarization} \label{sec:model}

In this section, we describe our Bayesian query-focused summarization
model (\bayesum).  This task is very similar to the standard ad-hoc IR
task, with the important distinction that we are comparing query
models against \emph{sentence models}, rather than against document
models.  The shortness of sentences means that one must do a good job
of creating the query models.

To maintain generality, so that our model is applicable to any problem
for which multiple relevant documents are known for a query, we
formulate our model in terms of \emph{relevance judgments}.  For a
collection of $D$ documents and $Q$ queries, we assume we have a $D
\times Q$ binary matrix $r$, where $r_{dq} = 1$ if an only if document
$d$ is relevant to query $q$.  In multidocument summarization,
$r_{dq}$ will be $1$ exactly when $d$ is in the document set
corresponding to query $q$; in search-engine summarization, it will be
$1$ exactly when $d$ is returned by the search engine for query $q$.

\subsection{Language Modeling for IR} \label{sec:IR}

\bayesum\ is built on the concept of language models for information
retrieval.  The idea behind the language modeling techniques used in
IR is to represent either queries or documents (or both) as
probability distributions, and then use standard probabilistic
techniques for comparing them.  These probability distributions are
almost always ``bag of words'' distributions that assign a probability
to words from a fixed vocabulary $\cV$.

One approach is to build a probability distribution for a given
document, $p_d(\cdot)$, and to look at the probability of a query
under that distribution: $p_d(q)$.  Documents are ranked according to
how \emph{likely} they make the query \cite{ponte98lm}.  Other
researchers have built probability distributions over queries
$p_q(\cdot)$ and ranked documents according to how likely they look
under the query model: $p_q(d)$ \cite{lafferty01lm}.  A third approach
builds a probability distribution $p_q(\cdot)$ for the query, a
probability distribution $p_d(\cdot)$ for the document and then
measures the \emph{similarity} between these two distributions using
KL divergence \cite{lavrenko02relevance}:

\begin{equation} \label{eq:kl}
\textit{KL}\left( p_q~\vert\vert~p_d \right) =
  \sum_{w \in \cV} p_q(w) \log \frac {p_q(w)} {p_d(w)}
\end{equation}

The KL divergence between two probability distributions is zero when
they are identical and otherwise strictly positive.  It implicitly
assumes that both distributions $p_q$ and $p_d$ have the same
\emph{support}: they assign non-zero probability to exactly the same
subset of $\cV$; in order to account for this, the distributions $p_q$
and $p_d$ are smoothed against a background general English model.
This final mode---the KL model---is the one on which \bayesum\ is
based.



\subsection{Bayesian Statistical Model}

In the language of information retrieval, the query-focused sentence
extraction task boils down to estimating a good \emph{query model},
$p_q(\cdot)$.  Once we have such a model, we could estimate sentence
models for each sentence in a relevant document, and rank the
sentences according to Eq~\eqref{eq:kl}.

The \bayesum\ system is based on the following model: we hypothesize
that a sentence appears in a document because it is relevant to some
query, because it provides background information about the document
(but is not relevant to a known query) or simply because it contains
useless, general English filler.  Similarly, we model each word as
appearing for one of those purposes.  More specifically, our model
assumes that each word can be assigned a discrete, exact source, such
as ``this word is relevant to query $q_1$'' or ``this word is general
English.''  At the sentence level, however, sentences are assigned
\emph{degrees}: ``this sentence is 60\% about query $q_1$, 30\%
background document information, and 10\% general English.''

To model this, we define a general English language model,
$p^G(\cdot)$ to capture the English filler.  Furthermore, for each
document $d_k$, we define a background document language model,
$p^{d_k}(\cdot)$; similarly, for each query $q_j$, we define a
query-specific language model $p^{q_j}(\cdot)$.  Every word in a
document $d_k$ is modeled as being generated from a mixture of $p^G$,
$p^{d_k}$ and $\{ p^{q_j} : \text{query }q_j\text{ is relevant to
document }d_k \}$.  Supposing there are $J$ total queries and $K$
total documents, we say that the $n$th word from the $s$th sentence in
document $d$, $w_{dsn}$, has a corresponding \emph{hidden variable},
$z_{dsn}$ that specifies exactly which of these distributions is used
to generate that one word.  In particular, $z_{dsn}$ is a vector of
length $1+J+K$, where exactly one element is $1$ and the rest are $0$.

At the sentence level, we introduce a second layer of hidden
variables.  For the $s$th sentence in document $d$, we let $\pi_{ds}$
be a vector also of length $1+J+K$ that represents our degree of
belief that this sentence came from any of the models.  The
$\pi_{ds}$s lie in the $J+K$-dimensional simplex $\De^{J+K} = \{ \vec
\th = \langle \th_1, \dots, \th_{J+K+1} \rangle : (\forall i)~\th_i
\geq 0, \sum_i \th_i = 1 \}$.  The interpretation of the $\pi$
variables is that if the ``general English'' component of $\pi$ is
$0.9$, then 90\% of the words in this sentence will be general
English.  The $\pi$ and $z$ variables are constrained so that a
sentence cannot be generated by a document language model other than
its own document and cannot be generated by a query language model for
a query to which it is not relevant.

Since the $\pi$s are unknown, and it is unlikely that there is a
``true'' correct value, we place a corpus-level prior on them.  Since
$\pi$ is a multinomial distribution over its corresponding $z$s, it is
natural to use a Dirichlet distribution as a prior over $\pi$.  A
Dirichlet distribution is parameterized by a vector $\al$ of equal
length to the corresponding multinomial parameter, again with the
positivity restriction, but no longer required to sum to one.  It has
continuous density over a variable $\th_1, \dots, \th_I$ given by:
$\Dir(\vec \th \| \vec \al) =
  \frac {\Ga\left(\sum_i \al_i\right)}
        {\prod_i \Ga(\al_i)}
  \prod_i \th_i^{\al_i-1}$.
The first term is a normalization term that ensures that $\int_{\De^I}
\ud \vec \th ~\Dir(\vec \th \| \vec \al) = 1$.

\subsection{Generative Story}

The generative story for our model defines a distribution over a
corpus of queries, $\{ q_j \}_{1:J}$, and documents, $\{ d_k
\}_{1:K}$, as follows:

\begin{enumerate}
\pitem For each query $j=1\dots J$:
Generate each word $q_{jn}$ in $q_j$ by $p^{q_j}(q_{jn})$
\pitem For each document $k=1\dots K$ and each sentence $s$ in
document $k$:
\begin{enumerate}
\pitem Select the current sentence degree $\pi_{ks}$ by $\Dir(\pi_{ks} \| \al) r_k(\pi_{ks})$
\pitem For each word $w_{ksn}$ in sentence $s$:
\begin{itemize}
\pitem Select the word source $z_{ksn}$ according to $\Mult(z \|
\pi_{ks})$
\pitem Generate the word $w_{ksn}$ by 
\pinch\pinch
\begin{equation*}
\brack{
p^G(w_{ksn})            & \text{if } z_{ksn} = 0 \\
p^{d_k}(w_{ksn})        & \text{if } z_{ksn} = k+1 \\
p^{q_j}(w_{ksn})        & \text{if } z_{ksn} = j+K+1
}
\end{equation*}
\end{itemize}
\end{enumerate}
\end{enumerate}

We used $r$ to denote relevance judgments: $r_k(\pi) = 0$ if any
document component of $\pi$ \emph{except} the one corresponding to $k$
is non-zero, or if any query component of $\pi$ \emph{except} those
queries to which document $k$ is deemed relevant is non-zero (this
prevents a document using the ``wrong'' document or query components).
We have further assumed that the $z$ vector is laid out so that $z_0$
corresponds to general English, $z_{k+1}$ corresponds to document
$d_k$ for $0 \leq j < J$ and that $z_{j+K+1}$ corresponds to query
$q_j$ for $0 \leq k < K$.

\subsection{Graphical Model}

\begin{figure}[t]
\hspace{1cm}
\psfig{figure=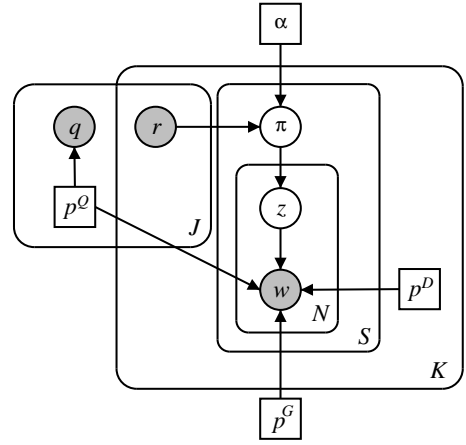,width=6cm}
\caption{Graphical model for the Bayesian Query-Focused Summarization Model.}
\label{fig:qfs-gm}
\end{figure}

The graphical model corresponding to this generative story is in
Figure~\ref{fig:qfs-gm}.  This model depicts the four known parameters
in square boxes ($\al$, $p^Q$, $p^D$ and $p^G$) with the three
observed random variables in shaded circles (the queries $q$, the
relevance judgments $r$ and the words $w$) and two unobserved random
variables in empty circles (the word-level indicator variables $z$ and
the sentence level degrees $\pi$).  The rounded plates denote
replication: there are $J$ queries and $K$ documents, containing $S$
sentences in a given document and $N$ words in a given sentence.  The
joint probability over the observed random variables is given in
Eq~\eqref{eq:full-joint}:

\pinch\pinch\pinch
\begin{align} \label{eq:full-joint}
&\p{\vec q_{1:J},r,\vec d_{1:K}} =
\Bigg[
  \prod_j
  \prod_n
    p^{q_j}\left(q_{jn}\right)
\Bigg] \times \\
&~~~~~~
\Bigg[
  \prod_k
  \prod_s
  \int_\De \ud \pi_{ks}~
    \p{\pi_{ks} \| \al, r} \nonumber\\
&~~~~~~~~~~~~~~~~
    \prod_n
      \sum_{z_{ksn}}
        \p{z_{ksn} \| \pi_{ks}}
        \p{w_{ksn} \| z_{ksn}}
\Bigg]
\nonumber
\end{align}
\pinch\pinch\pinch


This expression computes the probability of the data by integrating
out the unknown variables.  In the case of the $\pi$ variables, this
is accomplished by integrating over $\De$, the multinomial simplex,
according to the prior distribution given by $\al$.  In the case of
the $z$ variables, this is accomplished by summing over all possible
(discrete) values.  The final word probability is conditioned on the
$z$ value by selecting the appropriate distribution from $p^G$, $p^D$
and $p^Q$.  Computing this expression and finding optimal model
parameters is intractable due to the coupling of the variables under
the integral.

\section{Statistical Inference in \bayesum} \label{sec:inference}

Bayesian inference problems often give rise to intractable integrals,
and a large variety of techniques have been proposed to deal with
this.  The most popular are Markov Chain Monte Carlo (MCMC), the
Laplace (or saddle-point) approximation and the variational
approximation.  A third, less common, but very effective technique,
especially for dealing with mixture models, is expectation propagation
\cite{minka01thesis}.  In this paper, we will focus on expectation
propagation; experiments not reported here have shown variational EM
to perform comparably but take roughly 50\% longer to converge.

Expectation propagation (EP) is an inference technique introduced by
\newcite{minka01thesis} as a generalization of both belief propagation
and assumed density filtering.  In his thesis, Minka showed that EP is
very effective in mixture modeling problems, and later demonstrated
its superiority to variational techniques in the Generative Aspect
Model \cite{minka03gam}.  The key idea is to compute an integral of a
product of terms by iteratively applying a sequence of
``deletion/inclusion'' steps.  Given an integral of the form:
$\int_\De \ud \vec \pi~ p(\pi) \prod_n t_n(\vec \pi)$, EP approximates
each term $t_n$ by a simpler term $\tilde t_n$, giving Eq~\eqref{eq:ep}.

\pinch\pinch\pinch
\begin{equation} \label{eq:ep}
\int_\De \ud\vec\pi~ q(\vec \pi) \quad\quad
q(\vec \pi) = p(\vec \pi) \prod_n \tilde t_n(\vec \pi)
\end{equation}
\pinch\pinch\pinch

In each deletion/inclusion step, one of the approximate terms is
deleted from $q(\cdot)$, leaving $q^{-n}(\cdot) = q(\cdot) / \tilde
t_n(\cdot)$.  A new approximation for $t_n(\cdot)$ is computed so that
$t_n(\cdot)q^{-n}(\cdot)$ has the same integral, mean and variance as
$\tilde t_n(\cdot)q^{-n}(\cdot)$.  This new approximation, $\tilde
t_n(\cdot)$ is then included back into the full expression for
$q(\cdot)$ and the process repeats.  This algorithm always has a fixed
point and there are methods for ensuring that the approximation
remains in a location where the integral is well-defined.  Unlike
variational EM, the approximation given by EP is global, and often
leads to much more reliable estimates of the true integral.

In the case of our model, we follow \newcite{minka03gam}, who adapts
latent Dirichlet allocation of \newcite{blei-ng-jordan03lda} to EP.
Due to space constraints, we omit the inference algorithms and instead
direct the interested reader to the description given by
\newcite{minka03gam}.

\section{Search-Engine Experiments}

The first experiments we run are for query-focused single document
summarization, where relevant documents are returned from a search
engine, and a short summary is desired of each document.

\subsection{Data}

The data we use to train and test \bayesum\ is drawn from the Text
REtrieval Conference (TREC) competitions.  This data set consists of
queries, documents and relevance judgments, exactly as required by our
model.  The queries are typically broken down into four fields of
increasing length: the title (3-4 words), the summary (1 sentence),
the narrative (2-4 sentences) and the concepts (a list of keywords).
Obviously, one would expect that the longer the query, the better a
model would be able to do, and this is borne out experimentally
(Section~\ref{sec:eval-varying}).

Of the TREC data, we have trained our model on 350 queries (queries
numbered 51-350 and 401-450) and all corresponding relevant documents.
This amounts to roughly $43k$ documents, $2.1m$ sentences and $65.8m$
words.  The mean number of relevant documents per query is $137$ and
the median is $81$ (the most prolific query has $968$ relevant
documents).  On the other hand, each document is relevant to, on
average, $1.11$ queries (the median is $5.5$ and the most generally
relevant document is relevant to $20$ different queries).  In all
cases, we apply stemming using the Porter stemmer; for all other
models, we remove stop words.

In order to \emph{evaluate} our model, we had seven human judges
manually perform the query-focused sentence extraction task.  The
judges were supplied with the full TREC query and a single document
relevant to that query, and were asked to select up to four sentences
from the document that best met the needs given by the query.  Each
judge annotated $25$ queries with some overlap to allow for an
evaluation of inter-annotator agreement, yielding annotations for a
total of $166$ unique query/document pairs.  On the doubly annotated
data, we computed the inter-annotator agreement using the kappa measure.
The kappa value found was $0.58$, which is low, but not abysmal (also,
keep in mind that this is computed over only $25$ of the $166$
examples).  

\subsection{Evaluation Criteria}

Since there are differing numbers of sentences selected per document
by the human judges, one cannot compute precision and recall; instead,
we opt for other standard IR performance measures.  We consider three
related criteria: mean average precision (MAP), mean reciprocal rank
(MRR) and precision at 2 (P@2).  MAP is computed by calculating
precision at every sentence as ordered by the system up until all
relevant sentences are selected and averaged.  MRR is the reciprocal
of the rank of the first relevant sentence.  P@2 is the precision
computed at the first point that two relevant sentences have been
selected (in the rare case that humans selected only one sentence, we
use P@1).

\subsection{Baseline Models} \label{sec:baselines}

As baselines, we consider four strawman models and two
state-of-the-art information retrieval models.  The first strawman,
\textsc{Random} ranks sentences randomly.  The second strawman,
\textsc{Position}, ranks sentences according to their absolute
position (in the context of non-query-focused summarization, this is
an incredibly powerful baseline).  The third and fourth models are
based on the vector space interpretation of IR.  The third model,
\textsc{Jaccard}, uses standard Jaccard distance score (intersection
over union) between each sentence and the query to rank sentences.
The fourth, \textsc{Cosine}, uses TF-IDF weighted cosine similarity.

The two state-of-the-art IR models used as comparative systems are
based on the language modeling framework described in
Section~\ref{sec:IR}.  These systems compute a language model for each
query and for each sentence in a document.  Sentences are then ranked
according to the KL divergence between the query model and the
sentence model, smoothed against a general model estimated from the
entire collection, as described in the case of document retrieval by
\newcite{lavrenko02relevance}.  This is the first system we compare
against, called \textsc{KL}.

The second true system, \textsc{KL+Rel} is based on augmenting the
\textsc{KL} system with blind relevance feedback (query expansion).
Specifically, we first run each query against the document set
returned by the relevance judgments and retrieve the top $n$
sentences.  We then expand the query by interpolating the original
query model with a query model estimated on these sentences.  This
serves as a method of query expansion.  We ran experiments ranging $n$
in $\{ 5, 10, 25, 50, 100 \}$ and the interpolation parameter $\la$ in
$\{ 0.2, 0.4, 0.6, 0.8 \}$ and used oracle selection (on MRR) to
choose the values that performed best (the results are thus overly
optimistic).  These values were $n=25$ and $\la = 0.4$.

Of all the systems compared, only \bayesum\ and the \textsc{KL+Rel}
model use the relevance judgments; however, they both have access to
exactly the same information.  The other models only run on the subset
of the data used for evaluation (the corpus language model for the
\textsc{KL} system and the IDF values for the \textsc{Cosine} model
are computed on the full data set).  EP ran for 2.5 hours.

\subsection{Performance on all Query Fields}

\begin{table}[t]
\center
\begin{tabular}{|l|c|c|c|}
\hline
& {\bf MAP} & {\bf MRR} & {\bf P@2} \\
\hline
\textsc{Random}      & 19.9 & 37.3 & 16.6 \\
\textsc{Position}    & 24.8 & 41.6 & 19.9 \\
\hline
\textsc{Jaccard}     & 17.9 & 29.3 & 16.7 \\
\textsc{Cosine}      & 29.6 & 50.3 & 23.7 \\
\textsc{KL}          & 36.6 & 64.1 & 27.6 \\
\textsc{KL+Rel}      & 36.3 & 62.9 & 29.2 \\
\hline
\textsc{\bayesum}        & 44.1 & 70.8 & 33.6 \\
\hline
\end{tabular}
\caption{Empirical results for the baseline models as well as
  \bayesum, when all query fields are used.}
\label{tab:results-all}
\end{table}

Our first evaluation compares results when all query fields are used
(title, summary, description and concepts\footnote{A reviewer pointed
out that concepts were later removed from TREC because they were ``too
good.''  Section~\ref{sec:eval-varying} considers the case without the
concepts field.}).  These results are shown in
Table~\ref{tab:results-all}.  As we can see from these results, the
\textsc{Jaccard} system alone is not sufficient to beat the
position-based baseline.  The \textsc{Cosine} does beat the position
baseline by a bit of a margin (5 points better in MAP, 9 points in MRR
and 4 points in P@2), and is in turn beaten by the \textsc{KL} system
(which is 7 points, 14 points and 4 points better in MAP, MRR and P@2,
respectively).  Blind relevance feedback (parameters of which were
chosen by an oracle to maximize the P@2 metric) actually hurts MAP and
MRR performance by 0.3 and 1.2, respectively, and increases P@2 by
1.5.  Over the best performing baseline system (either \textsc{KL} or
\textsc{KL+Rel}), \bayesum\ wins by a margin of 7.5 points in MAP, 6.7
for MRR and 4.4 for P@2.

\subsection{Varying Query Fields} \label{sec:eval-varying}

Our next experimental comparison has to do with reducing the amount of
information given in the query.  In Table~\ref{tab:results-fields}, we
show the performance of the \textsc{KL}, \textsc{KL-Rel} and
\textsc{\bayesum} systems, as we use different query fields.  There are
several things to notice in these results.  First, the standard KL
model without blind relevance feedback performs worse than the
position-based model when only the 3-4 word title is available.
Second, \bayesum\ using \emph{only the title} outperform the KL
model \emph{with} relevance feedback using \emph{all fields}.  In
fact, one can apply \bayesum\ without using \emph{any} of the
query fields; in this case, only the relevance judgments are available
to make sense of what the query might be.  Even in this circumstance,
\bayesum\ achieves a MAP of $39.4$, an MRR of $64.7$ and a P@2 of
$30.4$, still better across the board than \textsc{KL-Rel} with all
query fields.  While initially this seems counterintuitive, it is
actually not so unreasonable: there is significantly more information
available in several hundred positive relevance judgments than in a
few sentences.  However, the simple blind relevance feedback mechanism
so popular in IR is unable to adequately model this.

\begin{table}[t]
\small \center
\begin{tabular}{|ll|c|c|c|}
\hline
&& {\bf MAP} & {\bf MRR} & {\bf P@2} \\
\hline
\multicolumn{2}{|l|}{\textsc{Position}}  
                     & 24.8 & 41.6 & 19.9 \\
\hline
Title
 & KL     & 19.9 & 32.6 & 17.8 \\
 & KL-Rel & 31.9 & 53.8 & 26.1 \\
 & \bayesum   & 41.1 & 65.7 & 31.6 \\
\hline
+Description
 & KL     & 31.5 & 58.3 & 24.1 \\
 & KL-Rel & 32.6 & 55.0 & 26.2 \\
 & \bayesum   & 40.9 & 66.9 & 31.0 \\
\hline
+Summary
 & KL     & 31.6 & 56.9 & 23.8 \\
 & KL-Rel & 34.2 & 48.5 & 27.0 \\
 & \bayesum   & 42.0 & 67.8 & 31.8 \\
\hline
+Concepts
 & KL     & 36.7 & 64.2 & 27.6 \\
 & KL-Rel & 36.3 & 62.9 & 29.2 \\
 & \bayesum   & 44.1 & 70.8 & 33.6 \\
\hline
\emph{No Query}
 & \bayesum   & 39.4 & 64.7 & 30.4 \\
\hline        
\end{tabular}
\caption{Empirical results for the position-based model, the KL-based
  models and \bayesum, with different inputs.}
\label{tab:results-fields}
\end{table}

With the exception of the KL model without relevance feedback, adding
the description on top of the title does not seem to make any
difference for any of the models (and, in fact, occasionally hurts
according to some metrics).  Adding the summary improves performance
in most cases, but not significantly.  Adding concepts tends to
improve results slightly more substantially than any other.

\subsection{Noisy Relevance Judgments} \label{sec:experiments:norel}

Our model hinges on the assumption that, for a given query, we have
access to a collection of \emph{known relevant documents.}  In most
real-world cases, this assumption is violated.  Even in multidocument
summarization as run in the DUC competitions, the assumption of access
to a collection of documents all relevant to a user need is
unrealistic.  In the real world, we will have to deal with document
collections that ``accidentally'' contain irrelevant documents.  The
experiments in this section show that \bayesum\ is comparatively
robust.

\begin{figure}[t]
\psfig{figure=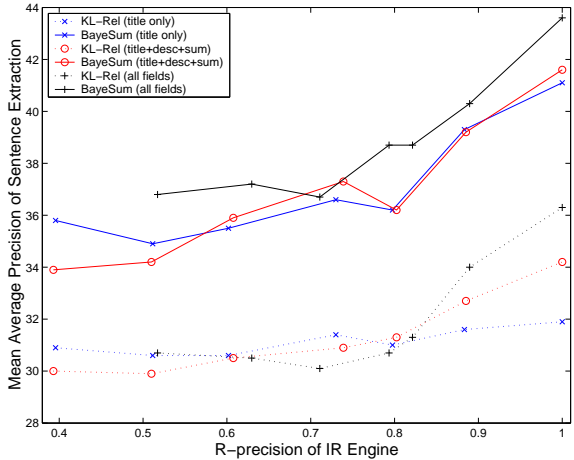,width=7.6cm}
\caption{Performance with noisy relevance judgments.  The X-axis is
  the R-precision of the IR engine and the Y-axis is the summarization
  performance in MAP.  Solid lines are \bayesum, dotted lines are KL-Rel.
  Blue/stars indicate title only, red/circles indicated
  title+description+summary and black/pluses indicate all fields.}
\label{fig:noisy}
\end{figure}

For this experiment, we use the IR engine that performed best in the
TREC 1 evaluation: Inquery \cite{callan92inquery}.  We used the
official TREC results of Inquery on the subset of the TREC corpus we
consider.  The Inquery R-precision on this task is $0.39$ using title
only, and $0.51$ using all fields.  In order to obtain curves as the
IR engine improves, we have linearly interpolated the Inquery rankings
with the true relevance judgments.  By tweaking the interpolation
parameter, we obtain an IR engine with improving performance, but with
a reasonable bias.  We have run both \bayesum\ and KL-Rel on the relevance
judgments obtained by this method for six values of the
interpolation parameter.  The results are shown in
Figure~\ref{fig:noisy}.

As we can observe from the figure, the solid lines (\bayesum) are
always above the dotted lines (KL-Rel).  Considering the KL-Rel
results alone, we can see that for a non-perfect IR engine, it makes
little difference what query fields we use for the summarization task:
they all obtain roughly equal scores.  This is because the performance
in KL-Rel is dominated by the performance of the IR engine.  Looking
only at the \bayesum\ results, we can see a much stronger, and perhaps
surprising difference.  For an imperfect IR system, it is better to
use only the title than to use the title, description and summary for
the summarization component.  We believe this is because the title is
more on topic than the other fields, which contain terms like ``A
relevant document should describe \dots.''  Nevertheless, \bayesum\
has a more upward trend than KL-Rel, which indicates that improved IR
will result in improved summarization for \bayesum\ but not for
KL-Rel.

\section{Multidocument Experiments}

We present two results using \bayesum\ in the multidocument
summarization settings, based on the official results from the
Multilingual Summarization Evaluation (MSE) and Document Understanding
Conference (DUC) competitions in 2005.

\subsection{Performance at MSE 2005}

We participated in the Multilingual Summarization Evaluation (MSE)
workshop with a system based on \bayesum.  The task for this
competition was generic (no query) multidocument summarization.
Fortunately, not having a query is not a hindrance to our model.  To
account for the redundancy present in document collections, we applied
a greedy selection technique that selects sentences central to the
document cluster but far from previously selected sentences
\cite{daume05mse}.  In MSE, our system performed very well.  According
to the human ``pyramid'' evaluation, our system came first with a
score of $0.529$; the next best score was $0.489$.  In the automatic
``Basic Element'' evaluation, our system scored $0.0704$ (with a 95\%
confidence interval of $[0.0429,0.1057]$), which was the third best
score on a site basis (out of 10 sites), and was not statistically
significantly different from the best system, which scored $0.0981$.

\subsection{Performance at DUC 2005}

We also participated in the Document Understanding Conference (DUC)
competition.  The chosen task for DUC was query-focused multidocument
summarization.  We entered a nearly identical system to DUC as to MSE,
with an additional rule-based sentence compression component
\cite{daume05duc}.  Human evaluators considered both
\emph{responsiveness} (how well did the summary answer the query) and
\emph{linguistic quality}.  Our system achieved the highest
responsiveness score in the competition.  We scored more poorly on the
linguistic quality evaluation, which (only 5 out of about 30 systems
performed worse); this is likely due to the sentence compression we
performed on top of \bayesum.  On the automatic Rouge-based
evaluations, our system performed between third and sixth (depending
on the Rouge parameters), but was never statistically significantly
worse than the best performing systems.

\section{Discussion and Future Work} \label{sec:discussion}

In this paper we have described a model for automatically generating a
query-focused summary, when one has access to multiple relevance
judgments.  Our Bayesian Query-Focused Summarization model (\bayesum)
consistently outperforms contending, state of the art information
retrieval models, even when it is forced to work with significantly
less information (either in the complexity of the query terms or the
quality of relevance judgments documents).  When we applied our system
as a stand-alone summarization model in the 2005 MSE and DUC tasks, we
achieved among the highest scores in the evaluation metrics.  The
primary weakness of the model is that it currently only operates in a
purely extractive setting.

One question that arises is: why does \bayesum\ so strongly outperform
KL-Rel, given that \bayesum\ can be seen as Bayesian formalism for
relevance feedback (query expansion)?  Both models have access to
exactly the same information: the queries and the true relevance
judgments.  This is especially interesting due to the fact that the
two relevance feedback parameters for KL-Rel were chosen
\emph{optimally} in our experiments, yet \bayesum\ consistently won
out.  One explanation for this performance win is that \bayesum\
provides a separate weight for each word, for each query.  This gives
it significantly more flexibility.  Doing something similar with
ad-hoc query expansion techniques is difficult due to the enormous
number of parameters; see, for instance, \cite{buckley95optimization}.

One significant advantage of working in the Bayesian statistical
framework is that it gives us a straightforward way to integrate other
sources of knowledge into our model in a coherent manner.  One could
consider, for instance, to extend this model to the multi-document
setting, where one would need to explicitly model redundancy across
documents.  Alternatively, one could include user models to account
for novelty or user preferences along the lines of
\newcite{zhang02filtering}.

Our model is similar in spirit to the random-walk summarization model
\cite{otterbacher05walks}.  However, our model has several advantages
over this technique.  First, our model has no tunable parameters: the
random-walk method has many (graph connectivity, various thresholds,
choice of similarity metrics, etc.).  Moreover, since our model is
properly Bayesian, it is straightforward to extend it to model other
aspects of the problem, or to related problems.  Doing so in a non
ad-hoc manner in the random-walk model would be nearly impossible.

Another interesting avenue of future work is to relax the bag-of-words
assumption.  Recent work has shown, in related models, how this can be
done for moving from bag-of-words models to bag-of-$n$gram models
\cite{wallach06beyond}; more interesting than moving to $n$grams would
be to move to dependency parse trees, which could likely be accounted
for in a similar fashion.  One could also potentially relax the
assumption that the relevance judgments are known, and attempt to
integrate them out as well, essentially simultaneously performing IR
and summarization.

\begin{small}
\paragraph{Acknowledgments.}
We thank Dave Blei and Tom Minka for discussions related to topic
models, and to the anonymous reviewers, whose comments have been of
great benefit.  This work was partially supported by the National
Science Foundation, Grant IIS-0326276.
\end{small}

\bibliography{bibfile}

\begin{thebibliography}{}

\bibitem[\protect\citename{Blei \bgroup et al.\egroup
  }2003]{blei-ng-jordan03lda}
David Blei, Andrew Ng, and Michael Jordan.
\newblock 2003.
\newblock Latent {D}irichlet allocation.
\newblock {\em Journal of Machine Learning Research (JMLR)}, 3:993--1022,
  January.

\bibitem[\protect\citename{Buckley and Salton}1995]{buckley95optimization}
Chris Buckley and Gerard Salton.
\newblock 1995.
\newblock Optimization of relevance feedback weights.
\newblock In {\em Proceedings of the Conference on Research and Developments in
  Information Retrieval (SIGIR)}.

\bibitem[\protect\citename{Callan \bgroup et al.\egroup }1992]{callan92inquery}
Jamie Callan, Bruce Croft, and Stephen Harding.
\newblock 1992.
\newblock The {INQUERY} retrieval system.
\newblock In {\em Proceedings of the 3rd International Conference on Database
  and Expert Systems Applications}.

\bibitem[\protect\citename{{Daum\'e III} and Marcu}2005a]{daume05mse}
Hal {Daum\'e III} and Daniel Marcu.
\newblock 2005a.
\newblock Bayesian multi-document summarization at {MSE}.
\newblock In {\em ACL 2005 Workshop on Intrinsic and Extrinsic Evaluation
  Measures}.

\bibitem[\protect\citename{{Daum\'e III} and Marcu}2005b]{daume05duc}
Hal {Daum\'e III} and Daniel Marcu.
\newblock 2005b.
\newblock Bayesian summarization at {DUC} and a suggestion for extrinsic
  evaluation.
\newblock In {\em Document Understanding Conference}.

\bibitem[\protect\citename{Lafferty and Zhai}2001]{lafferty01lm}
John Lafferty and {ChengXiang} Zhai.
\newblock 2001.
\newblock Document language models, query models, and risk minimization for
  information retrieval.
\newblock In {\em Proceedings of the Conference on Research and Developments in
  Information Retrieval (SIGIR)}.

\bibitem[\protect\citename{Lavrenko \bgroup et al.\egroup
  }2002]{lavrenko02relevance}
Victor Lavrenko, M.~Choquette, and Bruce Croft.
\newblock 2002.
\newblock Crosslingual relevance models.
\newblock In {\em Proceedings of the Conference on Research and Developments in
  Information Retrieval (SIGIR)}.

\bibitem[\protect\citename{Liu and Croft}2002]{liu02passage}
Xiaoyong Liu and Bruce Croft.
\newblock 2002.
\newblock Passage retrieval based on language models.
\newblock In {\em Processing of the Conference on Information and Knowledge
  Management (CIKM)}.

\bibitem[\protect\citename{Minka and Lafferty}2003]{minka03gam}
Thomas Minka and John Lafferty.
\newblock 2003.
\newblock Expectation-propagation for the generative aspect model.
\newblock In {\em Proceedings of the Converence on Uncertainty in Artificial
  Intelligence (UAI)}.

\bibitem[\protect\citename{Minka}2001]{minka01thesis}
Thomas Minka.
\newblock 2001.
\newblock {\em A family of algorithms for approximate {Bayesian} inference}.
\newblock {Ph.D.} thesis, Massachusetts Institute of Technology, Cambridge, MA.

\bibitem[\protect\citename{Murdock and Croft}2005]{murdock05sentence}
Vanessa Murdock and Bruce Croft.
\newblock 2005.
\newblock A translation model for sentence retrieval.
\newblock In {\em Proceedings of the Joint Conference on Human Language
  Technology Conference and Empirical Methods in Natural Language Processing
  (HLT/EMNLP)}, pages 684--691.

\bibitem[\protect\citename{Otterbacher \bgroup et al.\egroup
  }2005]{otterbacher05walks}
Jahna Otterbacher, Gunes Erkan, and Dragomir~R. Radev.
\newblock 2005.
\newblock Using random walks for question-focused sentence retrieval.
\newblock In {\em Proceedings of the Joint Conference on Human Language
  Technology Conference and Empirical Methods in Natural Language Processing
  (HLT/EMNLP)}.

\bibitem[\protect\citename{Ponte and Croft}1998]{ponte98lm}
Jay~M. Ponte and Bruce Croft.
\newblock 1998.
\newblock A language modeling approach to information retrieval.
\newblock In {\em Proceedings of the Conference on Research and Developments in
  Information Retrieval (SIGIR)}.

\bibitem[\protect\citename{Wallach}2006]{wallach06beyond}
Hanna Wallach.
\newblock 2006.
\newblock Topic modeling: beyond bag-of-words.
\newblock In {\em Proceedings of the International Conference on Machine
  Learning (ICML)}.

\bibitem[\protect\citename{Zhang \bgroup et al.\egroup }2002]{zhang02filtering}
Yi~Zhang, Jamie Callan, and Thomas Minka.
\newblock 2002.
\newblock Novelty and redundancy detection in adaptive filtering.
\newblock In {\em Proceedings of the Conference on Research and Developments in
  Information Retrieval (SIGIR)}.

\end{thebibliography}
\bibliographystyle{acl}

\end{document}